# DISCOUNTING AND COMBINATION OPERATIONS IN EVIDENTIAL REASONING

**J. W. Guan**
Department of Information Systems
University of Ulster at Jordanstown
Newtownabbey, Co. Antrim
BT37 0QB, Northern Ireland, U.K.
E-mail: CBFM23@UK.AC.ULSTER.UJVAX

**D. A. Bell**
Department of Information Systems
University of Ulster at Jordanstown
Newtownabbey, Co. Antrim
BT37 0QB, Northern Ireland, U.K.
E-mail: CBJW23@UK.AC.ULSTER.UPVAX

## Abstract

Evidential reasoning is now a leading topic in Artificial Intelligence. Evidence is represented by a variety of evidential functions. Evidential reasoning is carried out by certain kinds of fundamental operation on these functions. This paper discusses two of the basic operations on evidential functions, the discount operation and the well-known orthogonal sum operation. We show that the discount operation is not commutative with the orthogonal sum operation, and derive expressions for the two operations applied to the various evidential functions.

## 1 INTRODUCTION

In evidential reasoning, evidence is represented by evidential functions. Five such functions are commonly used: — mass function, belief function, plausibility function, commonality function, and doubt function. These functions are mutually transformable, conveying the same information, and a choice can be made between them to get the most appreciate one for a particular case.

Given a class of evidential functions $\mathcal{E}$, we can define various operations on it.

1. The *orthogonal sum* operation $\oplus$ carries out combination of evidence; it is known as Dempster-Shafer's rule.

2. The *discount* operation $\alpha$ on evidential functions reduces the mass assigned to the evidence by some proportion (Strat 1987).

There are three other operations, not considered in this paper. The refining operation $\sigma$ carries out detailing of evidence; the coarsening operation $\sigma^{-1}$ carries out summarizing of evidence; and the limit operation $lim$ on a sequence of evidential functions permits exploration of the most general evidential functions.

Evidential reasoning can be considered as an operational system (Guan & Bell 1993) with an evidential space $\mathcal{E}$ and five operations $\oplus, \sigma, \sigma^{-1}, \alpha, lim$, denoted by $< \mathcal{E}, \oplus, \sigma, \sigma^{-1}, \alpha, lim >$.

The pairwise commutativity of the operations varies.

This paper discusses the *discount and orthogonal sum* operations. We define the discount operation here, and show that it is not commutative with the orthogonal sum operation. This leads us to investigate the combination of discounted evidential functions further. Section 2 describes how to discount evidential functions. Section 3 discusses the relation between the discount operation $\alpha$ and the combination operation $\oplus$ on evidential functions. We show that these two basic operations are non-commutative. So we investigate the combination of discounted evidential functions further.

## 2 HOW TO DISCOUNT THE EVIDENTIAL FUNCTIONS

Let $m$ be a mass function on $2^\Theta$. Given a real number $\alpha$ such that $0 < \alpha < 1$, the function $m^\alpha$ defined by $m^\alpha(\Theta) = (1-\alpha)m(\Theta) + \alpha$ and $m^\alpha(A) = (1-\alpha)m(A)$ for all $A \subset \Theta$ is said to be the *discounted function with rate $\alpha$* of mass function $m$.

The idea is to " discount " $m(A)$ to $m^\alpha(A) = (1-\alpha)m(A)$ for all proper subsets $A$ of $\Theta, A \subset \Theta$.

As to at $\Theta$, notice that we must have

$$m^\alpha(\Theta) + \sum_{A \subset \Theta} m^\alpha(A) = 1,$$

$$m^\alpha(\Theta) = 1 - \sum_{A \subset \Theta} m^\alpha(A),$$

$$m^\alpha(\Theta) = 1 - \sum_{A \subset \Theta} (1-\alpha)m(A).$$

So we assign

$$1 - \sum_{A \subset \Theta} (1-\alpha)m(A)$$

to $m^\alpha(\Theta)$.

Recall that a mass function $m$ on $2^\Theta$ with positive value at $\Theta$: $m(\Theta) > 0$ is a support mass function (see section 12.3, p.70, Vol.2, Guan & Bell, 1992 — but notice that a support mass function on $2^\Theta$ may also have $m(\Theta) = 0$).

**THEOREM 1.** (*Discounting a mass function to a support mass function.*) Let $m$ be a mass function on $2^\Theta$. Then its discounted function $m^\alpha$ is a support mass function on $2^\Theta$ for all $0 < \alpha < 1$.

**PROOF.** We proceed as follows.

m1) We prove $m^\alpha(\emptyset) = 0$.

Indeed, we find that $m^\alpha(\emptyset) = (1-\alpha)m(\emptyset) = (1-\alpha) \times 0 = 0$ since $\emptyset \subset \Theta$.

m2) We prove $\sum_{X \subseteq \Theta} m^\alpha(X) = 1$.

Indeed, we find that

$$\sum_{X \subseteq \Theta} m^\alpha(X) = \sum_{X \subset \Theta} m^\alpha(X) + m^\alpha(\Theta)$$

$$= \sum_{X \subset \Theta} (1-\alpha)m(X) + ((1-\alpha)m(\Theta) + \alpha)$$

$$= (1-\alpha) \sum_{X \subset \Theta} m(X) + ((1-\alpha)m(\Theta) + \alpha)$$

$$= (1-\alpha)(1 - m(\Theta)) + (1-\alpha)m(\Theta) + \alpha$$

since $\sum_{X \subset \Theta} m(X) = 1 - m(\Theta)$ by the definition of mass function.

Thus, we have

$$\sum_{X \subseteq \Theta} m^\alpha(X) = (1-\alpha)(1 - m(\Theta)) + (1-\alpha)m(\Theta) + \alpha$$

$$= (1-\alpha) + \alpha = 1.$$

Finally, $m^\alpha$ is a support mass function on $2^\Theta$ since it is positive at $\Theta$:

$$m^\alpha(\Theta) = (1-\alpha)m(\Theta) + \alpha \geq \alpha > 0. \quad \boxed{\text{QED}}$$

In accordance with the usual pattern, the *discounted support* function, the *discounted support plausibility* function, the *discounted support commonality* function, and the *discounted support doubt* function of the *discounted support mass* function $m^\alpha$ are denoted by $bel^\alpha, pls^\alpha, com^\alpha, dou^\alpha$ and defined by

$$bel^\alpha(A) = \sum_{X \subseteq A} m^\alpha(X), pls^\alpha(A) = \sum_{X \cap A \neq \emptyset} m^\alpha(X),$$

$$com^\alpha(A) = \sum_{X \supseteq A} m^\alpha(X), dou^\alpha(A) = \sum_{X \subseteq \bar{A}} m^\alpha(X)$$

for all $A \subseteq \Theta$, respectively.

**THEOREM 2.** (*Discounting evidential functions.*) Let $m$ be a mass function on $2^\Theta$ and let $bel, pls, com, dou$ be its belief function, plausibility function, commonality function, and doubt function, respectively.

Let $m^\alpha$ be its discounted support mass function on $2^\Theta$ for $0 < \alpha < 1$ and let $bel^\alpha, pls^\alpha, com^\alpha, dou^\alpha$ be the belief function, the plausibility function, the commonality function, the doubt function of its discounted support mass function $m^\alpha$, respectively. Then

(1) For the discounted support function $bel^\alpha$ we have $bel^\alpha(A) = (1-\alpha)bel(A)$ for all $A \subset \Theta$, and of course $bel^\alpha(\Theta) = 1$ (since $bel^\alpha$ is a belief function on $2^\Theta$).

(2) For the discounted support plausibility function $pls^\alpha$ we have $pls^\alpha(A) = (1-\alpha)pls(A) + \alpha$ for all $A \neq \emptyset$, and of course $pls^\alpha(\emptyset) = 0$ (since $pls^\alpha$ is a plausibility function on $2^\Theta$).

(3) For the discounted support commonality function $com^\alpha$ we have $com^\alpha(A) = (1-\alpha)com(A) + \alpha$ for all $A \subseteq \Theta$.

(4) For the discounted support doubt function $dou^\alpha$ we have $dou^\alpha(A) = (1-\alpha)dou(A)$ for all $A \neq \emptyset$, and of course $dou^\alpha(\emptyset) = 1$ (since $dou^\alpha$ is a doubt function on $2^\Theta$).

**PROOF.** (1) $bel^\alpha(A) = \sum_{X \subseteq A} m^\alpha(X) = \sum_{X \subseteq A}(1-\alpha)m(X)$ for all $A \subset \Theta$ (because $X \subset \Theta$ since $X \subseteq A$, and thus $m^\alpha(X) = (1-\alpha)m(X)$ by the definition of $m^\alpha$),

$$= (1-\alpha) \sum_{X \subseteq A} m(X) = (1-\alpha)bel(A)$$

for all $A \subset \Theta$.

At $\Theta$ we find that $bel^\alpha(\Theta) = \sum_{X \subseteq \Theta} m^\alpha(X) = 1$ since we know that $m^\alpha$ is a mass function on $2^\Theta$ by theorem 1.

(2) $pls^\alpha(A) = 1 - bel^\alpha(\Theta - A)$

$= 1 - (1-\alpha)bel(\Theta - A)$ for all $A$ such that $\Theta - A \subset \Theta$ by (1) above.

That is, for all $A \neq \emptyset$,

$$pls^\alpha(A) = 1 - (1-\alpha)(1 - pls(A))$$

$$= 1 - (1-\alpha) + (1-\alpha)pls(A) = (1-\alpha)pls(A) + \alpha$$

for all $A \neq \emptyset$.

At $\emptyset$ we have $pls^\alpha(\emptyset) = 0$ since $pls^\alpha$ is a plausibility function on $2^\Theta$ (because $m^\alpha$ is a mass function by theorem 1).

(3) $com^\alpha(A) = \sum_{X \supseteq A} m^\alpha(X)$

$= \sum_{X \supseteq A, X \neq \Theta} m^\alpha(X) + m^\alpha(\Theta)$

$$= \sum_{X \supseteq A, X \neq \Theta} (1-\alpha)m(X) + (1-\alpha)m(\Theta) + \alpha$$

for all $A \subseteq \Theta$ by the definition $m^\alpha$,

$$= \sum_{X \supseteq A} (1-\alpha)m(X) + \alpha = (1-\alpha) \sum_{X \supseteq A} m(X) + \alpha$$

$$= (1-\alpha)com(A) + \alpha$$

for all $A \subseteq \Theta$.

$$(4) \quad dou^\alpha(A) = bel^\alpha(\Theta - A) = (1 - \alpha)bel(\Theta - A)$$

for all $A \supset \emptyset$ and so $\Theta - A \subset \Theta$,

$$= (1 - \alpha)dou(A)$$

for all $A \supset \emptyset$.

At $\emptyset$ we know that $dou^\alpha(\emptyset) = 1$ since $dou^\alpha$ is a doubt function on $2^\Theta$ (because $m^\alpha$ is a mass function on $2^\Theta$ by theorem 1.)   QED

Now, notice the following.

1. $m^{(\alpha_1\alpha_2)}$ is given by $m^{(\alpha_1\alpha_2)}(\Theta) = (1-\alpha_1\alpha_2)m(\Theta) + \alpha_1\alpha_2$, and $m^{(\alpha_1\alpha_2)}(A) = (1-\alpha_1\alpha_2)m(A)$ for all $A \subset \Theta$.

2. $(m^{\alpha_1})^{\alpha_2}$ is given by

$$(m^{\alpha_1})^{\alpha_2}(\Theta) = (1-\alpha_2)m^{\alpha_1}(\Theta) + \alpha_2$$
$$= (1-\alpha_2)((1-\alpha_1)m(\Theta) + \alpha_1) + \alpha_2$$
$$= (1-\alpha_2)(1-\alpha_1)m(\Theta) + (1-\alpha_2)\alpha_1 + \alpha_2$$
$$= (1-\alpha_2)(1-\alpha_1)m(\Theta) + \alpha_1 - \alpha_2\alpha_1 + \alpha_2,$$

and

$$(m^{\alpha_1})^{\alpha_2}(A) = (1-\alpha_2)m^{\alpha_1}(A) = (1-\alpha_2)((1-\alpha_1)m(A))$$
$$= (1-\alpha_2)(1-\alpha_1)m(A)$$

for all $A \subset \Theta$ since $m^{\alpha_1}(\Theta) = (1-\alpha_1)m(\Theta) + \alpha_1$, $m^{\alpha_1}(A) = (1-\alpha_1)m(A)$ for all $A \subset \Theta$.

3. $(m^{\alpha_2})^{\alpha_1}$ is given by

$$(m^{\alpha_2})^{\alpha_1}(\Theta) = (1-\alpha_1)m^{\alpha_2}(\Theta) + \alpha_1$$
$$= (1-\alpha_1)((1-\alpha_2)m(\Theta) + \alpha_2) + \alpha_1$$
$$= (1-\alpha_1)(1-\alpha_2)m(\Theta) + (1-\alpha_1)\alpha_2 + \alpha_1$$
$$= (1-\alpha_1)(1-\alpha_2)m(\Theta) + \alpha_2 - \alpha_1\alpha_2 + \alpha_1,$$

and

$$(m^{\alpha_2})^{\alpha_1}(A) = (1-\alpha_1)m^{\alpha_2}(A) = (1-\alpha_1)((1-\alpha_2)m(A))$$
$$= (1-\alpha_1)(1-\alpha_2)m(A)$$

for all $A \subset \Theta$ since $m^{\alpha_2}(\Theta) = (1-\alpha_2)m(\Theta) + \alpha_2$, $m^{\alpha_2}(A) = (1-\alpha_2)m(A)$ for all $A \subset \Theta$.

So we conclude the following:

(1) $m^{(\alpha_1\alpha_2)}$ is different from $(m^{\alpha_1})^{\alpha_2}$, and

(2) $(m^{\alpha_1})^{\alpha_2} = (m^{\alpha_2})^{\alpha_1}$.

## 3 DISCOUNTING AND COMBINATION OPERATIONS

We have these two basic operations on evidential functions. What is the relation between them? Are these two operations commutative?

That is, let $m_1, m_2$ be mass functions on $2^\Theta$ and let

$$bel_1, bel_2; pls_1, pls_2; com_1, com_2; dou_1, dou_2$$

be their belief functions, plausibility functions, commonality functions, and doubt functions, respectively. Do we have, for example, $(m_1 \oplus m_2)^\alpha = m_1^\alpha \oplus m_2^\alpha$? We establish that the answer in each case is " no ".

We have to look at each of these functions in detail, but we first give an example to demonstrate that the answer in each case is " no ". Then we derive formulae for the two orders of the operations, indicating what the generalization to an $n$-way orthogonal sum is in each case. This generalization can be established by induction, and we leave this as an exercise to the reader.

### 3.1  AN EXAMPLE FOR THE NON-COMMUTATIVITY OF DISCOUNT AND COMBINATION

**EXAMPLE 1.** Let $\Theta = \{a, b, c\}$ and let $m_1, m_2$ be mass functions on $2^\Theta$ as follows.

$$m_1(\{a,b\}) = 1/2, m_1(\Theta) = 1/2, m_1(elsewhere) = 0;$$
$$m_2(\{a,c\}) = 5/7, m_2(\Theta) = 2/7, m_2(elsewhere) = 0.$$

Also, let $\alpha = 0.3$.

Then, we proceed as follows.

(1) Mass functions — we show that $(m_1 \oplus m_2)^{0.3} \neq m_1^{0.3} \oplus m_2^{0.3}$.

(i) Compute $m_1 \oplus m_2$.

Make the intersection table to compute $m_1 \oplus m_2$.

| $m_1 \oplus m_2$ | $\{a,c\}5/7$ | $\Theta 2/7$ |
| --- | --- | --- |
| $\{a,b\}1/2$ | $\{a\}5/14$ | $\{a,b\}2/14$ |
| $\Theta 1/2$ | $\{a,c\}5/14$ | $\Theta 2/14$ |

We find that

$$(m_1 \oplus m_2)(\{a\}) = 5/14,$$
$$(m_1 \oplus m_2)(\{a,b\}) = 2/14 = 1/7,$$
$$(m_1 \oplus m_2)(\{a,c\}) = 5/14,$$
$$(m_1 \oplus m_2)(\Theta) = 2/14 = 1/7,$$
$$(m_1 \oplus m_2)(elsewhere) = 0.$$

(ii) Compute $(m_1 \oplus m_2)^{0.3}$:

$$(m_1 \oplus m_2)^{0.3}(\{a\}) = 0.7(5/14) = 0.25,$$
$$(m_1 \oplus m_2)^{0.3}(\{a,b\}) = 0.7(1/7) = 0.10,$$
$$(m_1 \oplus m_2)^{0.3}(\{a,c\}) = 0.7(5/14) = 0.25,$$
$$(m_1 \oplus m_2)^{0.3}(\Theta) = 0.7(1/7) + 0.3$$
$$= 0.10 + 0.3 = 0.40,$$
$$(m_1 \oplus m_2)^{0.3}(elsewhere) = 0.$$

(iii) Compute $m_1^{0.3} \oplus m_2^{0.3}$.

Compute $m_1^{0.3}$:

$$m_1^{0.3}(\{a,b\}) = 0.7 m_1(\{a,b\}) = 0.7(1/2) = 0.35,$$

$$m_1^{0.3}(\Theta) = 0.7 m_1(\Theta) + 0.3 = 0.7(1/2) + 0.3$$
$$= 0.35 + 0.3 = 0.65,$$
$$m_1^{0.3}(elsewhere) = 0.$$

Compute $m_2^{0.3}$:

$$m_2^{0.3}(\{a,c\}) = 0.7 m_1(\{a,c\}) = 0.7(5/7) = 0.5,$$
$$m_2^{0.3}(\Theta) = 0.7 m_1(\Theta) + 0.3 = 0.7(2/7) + 0.3$$
$$= 0.2 + 0.3 = 0.5,$$
$$m_2^{0.3}(elsewhere) = 0.$$

Now make the intersection table for $m_1^{0.3} \oplus m_2^{0.3}$.

| $m_1^{0.3} \oplus m_2^{0.3}$ | $\{a,c\}$ 0.5 | $\Theta$ 0.5 |
|---|---|---|
| $\{a,b\}$ 0.35 | $\{a\}$ 0.175 | $\{a,b\}$ 0.175 |
| $\Theta$ 0.65 | $\{a,c\}$ 0.325 | $\Theta$ 0.325 |

From the table, we find that $m_1^{0.3} \oplus m_2^{0.3}$ is the following.

$$(m_1^{0.3} \oplus m_2^{0.3})(\{a\}) = 0.175,$$
$$(m_1^{0.3} \oplus m_2^{0.3})(\{a,b\}) = 0.175,$$
$$(m_1^{0.3} \oplus m_2^{0.3})(\{a,c\}) = 0.325,$$
$$(m_1^{0.3} \oplus m_2^{0.3})(\Theta) = 0.325,$$
$$(m_1^{0.3} \oplus m_2^{0.3})(elsewhere) = 0.$$

By comparing (ii) and (iii) above we can see that $(m_1 \oplus m_2)^{0.3} \neq m_1^{0.3} \oplus m_2^{0.3}$.

(2) Belief functions — we show that $(bel_1 \oplus bel_2)^{0.3} \neq bel_1^{0.3} \oplus bel_2^{0.3}$.

(i) Compute $(bel_1 \oplus bel_2)^{0.3}$ by $(bel_1 \oplus bel_2)^{0.3}(A) = \sum_{X \subseteq A}(m_1 \oplus m_2)^{0.3}(X)$ for all $A \subseteq \Theta$.

By (1).(ii) above we find that

$$(bel_1 \oplus bel_2)^{0.3}(\emptyset) = 0, (bel_1 \oplus bel_2)^{0.3}(\Theta) = 1,$$
$$(bel_1 \oplus bel_2)^{0.3}(\{a\}) = (m_1 \oplus m_2)^{0.3}(\{a\}) = 0.25,$$
$$(bel_1 \oplus bel_2)^{0.3}(\{b\}) = (m_1 \oplus m_2)^{0.3}(\{b\}) = 0,$$
$$(bel_1 \oplus bel_2)^{0.3}(\{c\}) = (m_1 \oplus m_2)^{0.3}(\{c\}) = 0,$$
$$(bel_1 \oplus bel_2)^{0.3}(\{a,b\}) = (m_1 \oplus m_2)^{0.3}(\{a\})$$
$$+ (m_1 \oplus m_2)^{0.3}(\{b\}) + (m_1 \oplus m_2)^{0.3}(\{a,b\})$$
$$= 0.25 + 0 + 0.1 = 0.35,$$
$$(bel_1 \oplus bel_2)^{0.3}(\{b,c\}) = (m_1 \oplus m_2)^{0.3}(\{b\})$$
$$+ (m_1 \oplus m_2)^{0.3}(\{c\}) + (m_1 \oplus m_2)^{0.3}(\{b,c\})$$
$$= 0 + 0 + 0 = 0,$$
$$(bel_1 \oplus bel_2)^{0.3}(\{a,c\}) = (m_1 \oplus m_2)^{0.3}(\{a\})$$
$$+ (m_1 \oplus m_2)^{0.3}(\{c\}) + (m_1 \oplus m_2)^{0.3}(\{a,c\})$$
$$= 0.25 + 0 + 0.25 = 0.5.$$

(ii) Compute $bel_1^{0.3} \oplus bel_2^{0.3}$ by $(bel_1^{0.3} \oplus bel_2^{0.3})(A) = \sum_{X \subseteq A}(m_1^{0.3} \oplus m_2^{0.3})(X)$ for all $A \subseteq \Theta$.

By (1).(iii) we find that

$$(bel_1^{0.3} \oplus bel_2^{0.3})(\emptyset) = 0, (bel_1^{0.3} \oplus bel_2^{0.3})(\Theta) = 1,$$
$$(bel_1^{0.3} \oplus bel_2^{0.3})(\{a\}) = (m_1^{0.3} \oplus m_2^{0.3})(\{a\}) = 0.175,$$
$$(bel_1^{0.3} \oplus bel_2^{0.3})(\{b\}) = (m_1^{0.3} \oplus m_2^{0.3})(\{b\}) = 0,$$
$$(bel_1^{0.3} \oplus bel_2^{0.3})(\{c\}) = (m_1^{0.3} \oplus m_2^{0.3})(\{c\}) = 0,$$
$$(bel_1^{0.3} \oplus bel_2^{0.3})(\{a,b\}) = (m_1^{0.3} \oplus m_2^{0.3})(\{a\})$$
$$+ (m_1^{0.3} \oplus m_2^{0.3})(\{b\}) + (m_1^{0.3} \oplus m_2^{0.3})(\{a,b\})$$
$$= 0.175 + 0 + 0.175 = 0.35,$$
$$(bel_1^{0.3} \oplus bel_2^{0.3})(\{b,c\}) = (m_1^{0.3} \oplus m_2^{0.3})(\{b\})$$
$$+ (m_1^{0.3} \oplus m_2^{0.3})(\{c\}) + (m_1^{0.3} \oplus m_2^{0.3})(\{b,c\})$$
$$= 0 + 0 + 0 = 0,$$
$$(bel_1^{0.3} \oplus bel_2^{0.3})(\{a,c\}) = (m_1^{0.3} \oplus m_2^{0.3})(\{a\})$$
$$+ (m_1^{0.3} \oplus m_2^{0.3})(\{c\}) + (m_1^{0.3} \oplus m_2^{0.3})(\{a,c\})$$
$$= 0.175 + 0 + 0.325 = 0.5.$$

So we can see that $(bel_1 \oplus bel_2)^{0.3} \neq bel_1^{0.3} \oplus bel_2^{0.3}$ since $(bel_1 \oplus bel_2)^{0.3}(\{a\}) = 0.25$ in (i) above, but $(bel_1^{0.3} \oplus bel_2^{0.3})(\{a\}) = 0.175$ in (ii) above.

(3) We can also show that

$$(dou_1 \oplus dou_2)^{0.3} \neq dou_1^{0.3} \oplus dou_2^{0.3},$$
$$(pls_1 \oplus pls_2)^{0.3} \neq pls_1^{0.3} \oplus pls_2^{0.3},$$
$$(com_1 \oplus com_2)^{0.3} \neq com_1^{0.3} \oplus com_2^{0.3}. \diamond$$

### 3.2 THE FORMULAE FOR $(m_1 \oplus m_2)^\alpha$ AND $m_1^\alpha \oplus m_2^\alpha$

Now we start to establish these results formally.

**THEOREM 3.** (*Discounting an orthogonal sum of mass functions.*) Let $m_1, m_2$ be mass functions on $2^\Theta$ and let $0 < \alpha < 1$. Then we have the following formula for $(m_1 \oplus m_2)^\alpha$:

$$(m_1 \oplus m_2)^\alpha(\emptyset) = 0,$$
$$(m_1 \oplus m_2)^\alpha(A)$$
$$= (1-\alpha)\frac{\sum_{X \subset \Theta \text{ or } Y \subset \Theta, X \cap Y = A} m_1(X) m_2(Y)}{\sum_{X \cap Y \neq \emptyset} m_1(X) m_2(Y)}$$

for all $\emptyset \subset A \subset \Theta$; and at $\Theta$: $(m_1 \oplus m_2)^\alpha(\Theta) = (1-\alpha)\frac{m_1(\Theta) m_2(\Theta)}{\sum_{X \cap Y \neq \emptyset} m_1(X) m_2(Y)} + \alpha.$

**PROOF.** Remember that, by definition,

$$(m_1 \oplus m_2)(A) = \frac{\sum_{X \cap Y = A} m_1(X) m_2(Y)}{\sum_{X \cap Y \neq \emptyset} m_1(X) m_2(Y)}$$

for all $A \subseteq \Theta$.

Also, notice that

1. $X \cap Y \subset \Theta$ if and only if $X \subset \Theta$ or $Y \subset \Theta$.
2. $X \cap Y = \Theta$ if and only if $X = Y = \Theta$.

At $\emptyset$ we find that

$(m_1 \oplus m_2)^\alpha(\emptyset) = (1-\alpha)(m_1 \oplus m_2)(\emptyset) = (1-\alpha) \times 0 = 0.$

For all $A \subset \Theta, A \neq \emptyset$ we find that $(m_1 \oplus m_2)^\alpha(A) = (1-\alpha)(m_1 \oplus m_2)(A)$

$= (1-\alpha) \dfrac{\sum_{X \cap Y = A, A \subset \Theta} m_1(X) m_2(Y)}{\sum_{X \cap Y \neq \emptyset} m_1(X) m_2(Y)}$

$= (1-\alpha) \dfrac{\sum_{X \subset \Theta \text{ or } Y \subset \Theta, X \cap Y = A} m_1(X) m_2(Y)}{\sum_{X \cap Y \neq \emptyset} m_1(X) m_2(Y)}.$

At $\Theta$ we find that $(m_1 \oplus m_2)^\alpha(\Theta) = (1-\alpha)(m_1 \oplus m_2)(\Theta) + \alpha$

$= (1-\alpha) \dfrac{\sum_{X \cap Y = \Theta} m_1(X) m_2(Y)}{\sum_{X \cap Y \neq \emptyset} m_1(X) m_2(Y)} + \alpha$

$= (1-\alpha) \dfrac{\sum_{X = Y = \Theta} m_1(X) m_2(Y)}{\sum_{X \cap Y \neq \emptyset} m_1(X) m_2(Y)} + \alpha$

$= (1-\alpha) \dfrac{m_1(\Theta) m_2(\Theta)}{\sum_{X \cap Y \neq \emptyset} m_1(X) m_2(Y)} + \alpha.$ $\boxed{\text{QED}}$

More generally, for any finite number of operands, let $m_1, m_2, \ldots, m_n; n > 1$ be mass functions on $2^\Theta$ and let $0 < \alpha < 1$. Then we have the following formula for $(m_1 \oplus m_2 \oplus \ldots \oplus m_n)^\alpha$: $(m_1 \oplus m_2 \oplus \ldots \oplus m_n)^\alpha(\emptyset) = 0$,

$(m_1 \oplus m_2 \oplus \ldots \oplus m_n)^\alpha(A) = (1-\alpha) \times$

$\times ( \sum_{X_1 \subset \Theta \text{ or } \ldots \text{ or } X_n \subset \Theta, X_1 \cap X_2 \cap \ldots \cap X_n = A} m_1(X_1) m_2(X_2) \ldots m_n(X_n))$

$\div ( \sum_{X_1 \cap X_2 \cap \ldots \cap X_n \neq \emptyset} m_1(X_1) m_2(X_2) \ldots m_n(X_n))$

for all $\emptyset \subset A \subset \Theta$; and at $\Theta$: $(m_1 \oplus m_2 \oplus \ldots \oplus m_n)^\alpha(\Theta)$

$= (1-\alpha) \dfrac{m_1(\Theta) m_2(\Theta) \ldots m_n(\Theta)}{\sum_{X_1 \cap X_2 \cap \ldots \cap X_n \neq \emptyset} m_1(X_1) m_2(X_2) \ldots m_n(X)} + \alpha.$

**THEOREM 4.** (*Combining discounted mass functions.*) Let $m_1, m_2$ be mass functions on $2^\Theta$ and let $0 < \alpha < 1$. Then we have the following formula for $m_1^\alpha \oplus m_2^\alpha$: $(m_1^\alpha \oplus m_2^\alpha)(\emptyset) = 0$,

$(m_1^\alpha \oplus m_2^\alpha)(A)$

$= (1-\alpha)^2 \times ( \sum_{X \cap Y = A} m_1(X) m_2(Y))$

$+ \alpha(1-\alpha)(m_1(A) + m_2(A))$

$\div ((1-\alpha)^2 \sum_{X \cap Y \neq \emptyset} m_1(X) m_2(Y) + 2\alpha - \alpha^2)$

for all $\emptyset \subset A \subset \Theta$, and at $\Theta$

$(m_1^\alpha \oplus m_2^\alpha)(\Theta) = ((1-\alpha)^2 m_1(\Theta) m_2(\Theta)$

$+ \alpha(1-\alpha)(m_1(\Theta) + m_2(\Theta)) + \alpha^2)$

$\div ((1-\alpha)^2 \sum_{X \cap Y \neq \emptyset} m_1(X) m_2(Y) + 2\alpha - \alpha^2).$

**PROOF.** We know that $(m_1^\alpha \oplus m_2^\alpha)(\emptyset) = 0$, $(m_1^\alpha \oplus m_2^\alpha)(A) = (1/N_\alpha) \sum_{X \cap Y = A} m_1^\alpha(X) m_2^\alpha(Y)$ for all $A \subseteq \Theta, A \neq \emptyset$, where $N_\alpha = \sum_{X \cap Y \neq \emptyset} m_1^\alpha(X) m_2^\alpha(Y)$.

Also, again

1. $X \cap Y \subset \Theta$ if and only if $X \subset \Theta$ or $X \subset \Theta$.

2. $X \cap Y = \Theta$ if and only if $X = Y = \Theta$.

Then we find the following.

(1) For $N_\alpha$, we show that $N_\alpha = (1-\alpha)^2 N + 2\alpha - \alpha^2$, where $N = \sum_{X \cap Y \neq \emptyset} m_1(X) m_2(Y)$.

To show this, we proceed as follows. $N_\alpha = \sum_{X \cap Y \neq \emptyset} m_1^\alpha(X) m_2^\alpha(Y)$

$= \sum_{X = \Theta, Y = \Theta; (X \cap Y = \Theta)} m_1^\alpha(X) m_2^\alpha(Y)$

$+ \sum_{X = \Theta, \emptyset \subset Y \subset \Theta; (X \cap Y = Y)} m_1^\alpha(X) m_2^\alpha(Y)$

$+ \sum_{\emptyset \subset X \subset \Theta, Y = \Theta; (X \cap Y = X)} m_1^\alpha(X) m_2^\alpha(Y)$

$+ \sum_{\emptyset \subset X \subset \Theta, \emptyset \subset Y \subset \Theta; X \cap Y \neq \emptyset} m_1^\alpha(X) m_2^\alpha(Y)$

$= m_1^\alpha(\Theta) m_2^\alpha(\Theta) + m_1^\alpha(\Theta) \sum_{\emptyset \subset Y \subset \Theta} m_2^\alpha(Y)$

$+ m_2^\alpha(\Theta) \sum_{\emptyset \subset X \subset \Theta} m_1^\alpha(X)$

$+ \sum_{\emptyset \subset X \subset \Theta, \emptyset \subset Y \subset \Theta; X \cap Y \neq \emptyset} (1-\alpha) m_1(X)(1-\alpha) m_2(Y)$

$= ((1-\alpha) m_1(\Theta) + \alpha)((1-\alpha) m_2(\Theta) + \alpha)$

$+ ((1-\alpha) m_1(\Theta) + \alpha) \sum_{\emptyset \subset Y \subset \Theta} (1-\alpha) m_2(Y)$

$+ ((1-\alpha) m_2(\Theta) + \alpha) \sum_{\emptyset \subset X \subset \Theta} (1-\alpha) m_1(X)$

$+ \sum_{\emptyset \subset X \subset \Theta, \emptyset \subset Y \subset \Theta; X \cap Y \neq \emptyset} (1-\alpha) m_1(X)(1-\alpha) m_2(Y)$

$= (1-\alpha)^2 (m_1(\Theta) m_2(\Theta) + m_1(\Theta) \sum_{\emptyset \subset Y \subset \Theta} m_2(Y)$

$+ m_2(\Theta) \sum_{\emptyset \subset X \subset \Theta} m_1(X)$

$+ \sum_{\emptyset \subset X \subset \Theta, \emptyset \subset Y \subset \Theta; X \cap Y \neq \emptyset} m_1(X) m_2(Y))$

$+ \alpha^2 + \alpha(1-\alpha)(m_1(\Theta) + m_2(\Theta) + \sum_{\emptyset \subset Y \subset \Theta} m_2(Y)$

$+ \sum_{\emptyset \subset X \subset \Theta} m_1(X))$

$$= (1-\alpha)^2(\sum_{X=\Theta,Y=\Theta;(X\cap Y=\Theta)} m_1(X)m_2(Y)$$
$$+ \sum_{X=\Theta,\emptyset\subset Y\subset\Theta;(X\cap Y=Y)} m_1(X)m_2(Y)$$
$$+ \sum_{\emptyset\subset X\subset\Theta,Y=\Theta;(X\cap Y=X)} m_1(X)m_2(Y)$$
$$+ \sum_{\emptyset\subset X\subset\Theta,\emptyset\subset Y\subset\Theta;X\cap Y\neq\emptyset} m_1(X)m_2(Y))$$
$$+\alpha^2 + \alpha(1-\alpha)(\sum_{\emptyset\subset Y\subseteq\Theta} m_2(Y) + \sum_{\emptyset\subset X\subseteq\Theta} m_1(X))$$
$$= (1-\alpha)^2(\sum_{X\cap Y\neq\emptyset} m_1(X)m_2(Y)) + \alpha^2 + \alpha(1-\alpha)(1+1)$$
$$= (1-\alpha)^2 N + \alpha^2 + 2\alpha(1-\alpha) = (1-\alpha)^2 N + 2\alpha - \alpha^2.$$

That is, $N_\alpha = (1-\alpha)^2 N + 2\alpha - \alpha^2$, where $N = \sum_{X\cap Y\neq\emptyset} m_1(X)m_2(Y)$.

(2) We now show that at $\Theta$, $\sum_{X\cap Y=\Theta} m_1^\alpha(X)m_2^\alpha(Y)$
$= (1-\alpha)^2 m_1(\Theta)m_2(\Theta) + \alpha(1-\alpha)(m_1(\Theta)+m_1(\Theta)) + \alpha^2$.

We proceed as follows. $\sum_{X\cap Y=\Theta} m_1^\alpha(X)m_2^\alpha(Y) = m_1^\alpha(\Theta)m_2^\alpha(\Theta)$

$$= ((1-\alpha)m_1(\Theta) + \alpha)((1-\alpha)m_2(\Theta) + \alpha)$$
$$= (1-\alpha)^2 m_1(\Theta)m_2(\Theta) + \alpha(1-\alpha)(m_1(\Theta)+m_2(\Theta)) + \alpha^2$$

(3) For all $A \subset \Theta$ we establish that $\sum_{X\cap Y=A} m_1^\alpha(X)m_2^\alpha(Y)$

$$= (1-\alpha)^2(\sum_{X\cap Y=A} m_1(X)m_2(Y))$$
$$+ \alpha(1-\alpha)(m_1(A)+m_2(A)).$$

Indeed, we find that $\sum_{X\cap Y=A} m_1^\alpha(X)m_2^\alpha(Y) = \sum_{X=\Theta,Y=A;(X\cap Y=A)} m_1^\alpha(X)m_2^\alpha(Y)$

$$+ \sum_{X=A,Y=\Theta;(X\cap Y=A)} m_1^\alpha(X)m_2^\alpha(Y)$$
$$+ \sum_{X\subset\Theta,Y\subset\Theta,X\cap Y=A} m_1^\alpha(X)m_2^\alpha(Y)$$
$$= m_1^\alpha(\Theta)m_2^\alpha(A) + m_1^\alpha(A)m_2^\alpha(\Theta)$$
$$+ \sum_{X\subset\Theta,Y\subset\Theta,X\cap Y=A} m_1^\alpha(X)m_2^\alpha(Y)$$
$$= ((1-\alpha)m_1(\Theta) + \alpha)(1-\alpha)m_2(A)$$
$$+(1-\alpha)m_1(A)((1-\alpha)m_2(\Theta) + \alpha)$$
$$+ \sum_{X\subset\Theta,Y\subset\Theta,X\cap Y=A} (1-\alpha)^2 m_1(X)m_2(Y)$$
$$= (1-\alpha)^2(m_1(\Theta)m_2(A) + m_1(A)m_2(\Theta)$$
$$+ \sum_{X\subset\Theta,Y\subset\Theta,X\cap Y=A} m_1(X)m_2(Y))$$
$$+\alpha(1-\alpha)(m_2(A)+m_1(A))$$

$$= (1-\alpha)^2(\sum_{X=\Theta,Y=A;(X\cap Y=A)} m_1(X)m_2(Y)$$
$$+ \sum_{X=A,Y=\Theta;(X\cap Y=A)} m_1(X)m_2(Y)$$
$$+ \sum_{X\subset\Theta,Y\subset\Theta,X\cap Y=A} m_1(X)m_2(Y))$$
$$+\alpha(1-\alpha)(m_2(A)+m_1(A))$$
$$= (1-\alpha)^2(\sum_{X\cap Y=A} m_1(X)m_2(Y))$$
$$+\alpha(1-\alpha)(m_1(A)+m_2(A)). \quad \boxed{\text{QED}}$$

### 3.3 THE FORMULAE FOR $(bel_1 \oplus bel_2)^\alpha$ AND $bel_1^\alpha \oplus bel_2^\alpha$

In the next subsections, we establish similar results over our evidence space $\mathcal{E}$. First, belief functions.

**THEOREM 5.** (*Discounting an orthogonal sum of belief functions.*) Let $m_1, m_2$ be mass functions on $2^\Theta$ and let $bel_1, bel_2$ be their belief functions, respectively. Let $0 < \alpha < 1$. Then we have the following formula for $(bel_1 \oplus bel_2)^\alpha$:

$$(bel_1 \oplus bel_2)^\alpha(\emptyset) = 0, (bel_1 \oplus bel_2)^\alpha(\Theta) = 1,$$
$$(bel_1 \oplus bel_2)^\alpha(A)$$
$$= (1-\alpha)\frac{\sum_{X\subset\Theta \text{ or } Y\subset\Theta, X\cap Y\subseteq A} m_1(X)m_2(Y)}{\sum_{X\cap Y\neq\emptyset} m_1(X)m_2(Y)}$$

for all $\emptyset \subset A \subset \Theta$.

**PROOF.** By theorem 2.(1), we have $(bel_1 \oplus bel_2)^\alpha(\emptyset) = 0, (bel_1 \oplus bel_2)^\alpha(\Theta) = 1$.

For all $A \subset \Theta$, by definition we find that $(bel_1 \oplus bel_2)^\alpha(A) = \sum_{Z\subseteq A}(m_1 \oplus m_2)^\alpha(Z)$

$$= \sum_{Z\subseteq A}(1-\alpha)\frac{\sum_{X\subset\Theta \text{ or } Y\subset\Theta, X\cap Y=Z} m_1(X)m_2(Y)}{\sum_{X\cap Y\neq\emptyset} m_1(X)m_2(Y)}$$

since $Z \subseteq A$ and so $Z \subset \Theta$ and from theorem 3,

$$= (1-\alpha)\frac{\sum_{X\subset\Theta \text{ or } Y\subset\Theta, X\cap Y\subseteq A} m_1(X)m_2(Y)}{\sum_{X\cap Y\neq\emptyset} m_1(X)m_2(Y)}. \quad \boxed{\text{QED}}$$

More generally, let $m_1, m_2, ..., m_n; n > 1$ be mass functions on $2^\Theta$ and let $bel_1, bel_2, ..., bel_n$ be their belief functions, respectively. Let $0 < \alpha < 1$. Then we have the following formula for $(bel_1 \oplus bel_2 \oplus ... \oplus bel_n)^\alpha$:

$$(bel_1 \oplus bel_2 \oplus ... \oplus bel_n)^\alpha(\emptyset) = 0,$$
$$(bel_1 \oplus bel_2 \oplus ... \oplus bel_n)^\alpha(\Theta) = 1,$$
$$(bel_1 \oplus bel_2 \oplus ... \oplus bel_n)^\alpha(A) = (1-\alpha) \times$$
$$\times (\sum_{X_1\subset\Theta \text{ or } ... \text{ or } X_n\subset\Theta, X_1\cap X_2\cap...\cap X_n\subseteq A}$$
$$m_1(X_1)m_2(X_2)...m_n(X_n))$$

$$\div (\sum_{X_1 \cap X_2 \cap ... \cap X_n \neq \emptyset} m_1(X_1)m_2(X_2)...m_n(X_n))$$

for all $\emptyset \subset A \subset \Theta$.

**THEOREM 6.** (*Combining discounted belief functions.*) Let $m_1, m_2$ be mass functions on $2^\Theta$ and let $bel_1, bel_2$ be their belief functions, respectively. Let $0 < \alpha < 1$. Then we have the following formula for $bel_1^\alpha \oplus bel_2^\alpha$: $(bel_1^\alpha \oplus bel_2^\alpha)(\emptyset) = 0, (bel_1^\alpha \oplus bel_2^\alpha)(\Theta) = 1$, and $(bel_1^\alpha \oplus bel_2^\alpha)(A)$

$$= ((1-\alpha)^2 (\sum_{X \cap Y \subseteq A} m_1(X)m_2(Y))$$

$$+ \alpha(1-\alpha)(\sum_{Z \subseteq A}(m_1(Z) + m_2(Z))))$$

$$\div ((1-\alpha)^2 \sum_{X \cap Y \neq \emptyset} m_1(X)m_2(Y) + 2\alpha - \alpha^2)$$

for all $\emptyset \subset A \subset \Theta$.

**PROOF.** We know that $bel_1^\alpha \oplus bel_2^\alpha$ is a belief function. So of course we have

$$(bel_1^\alpha \oplus bel_2^\alpha)(\emptyset) = 0, (bel_1^\alpha \oplus bel_2^\alpha)(\Theta) = 1.$$

For all $A \subset \Theta$, by definition we find that $(bel_1^\alpha \oplus bel_2^\alpha)(A) = \sum_{Z \subseteq A}(m_1^\alpha \oplus m_2^\alpha)(Z)$

$$= \sum_{Z \subseteq A}((1-\alpha)^2 (\sum_{X \cap Y = Z} m_1(X)m_2(Y))$$

$$+ \alpha(1-\alpha)(m_1(Z) + m_2(Z)))$$

$$\div ((1-\alpha)^2 \sum_{X \cap Y \neq \emptyset} m_1(X)m_2(Y) + 2\alpha - \alpha^2)$$

since $Z \subseteq A$ and so $Z \subset \Theta$ and from theorem 4,

$$= ((1-\alpha)^2 (\sum_{X \cap Y \subseteq A} m_1(X)m_2(Y))$$

$$+ \alpha(1-\alpha)(\sum_{Z \subseteq A}(m_1(Z) + m_2(Z))))$$

$$\div ((1-\alpha)^2 \sum_{X \cap Y \neq \emptyset} m_1(X)m_2(Y) + 2\alpha - \alpha^2).$$

QED

### 3.4 THE FORMULAE FOR $(pls_1 \oplus pls_2)^\alpha$ AND $pls_1^\alpha \oplus pls_2^\alpha$

Now, plausibility functions.

**THEOREM 7.** (*Discounting an orthogonal sum of plausibility functions.*) Let $m_1, m_2$ be mass functions on $2^\Theta$ and let $pls_1, pls_2$ be their plausibility functions, respectively. Let $0 < \alpha < 1$. Then we have the following formula for $(pls_1 \oplus pls_2)^\alpha$: $(pls_1 \oplus pls_2)^\alpha(\emptyset) = 0, (pls_1 \oplus pls_2)^\alpha(\Theta) = 1$,

$$(pls_1 \oplus pls_2)^\alpha(A)$$

$$= 1 - (1-\alpha)\frac{\sum_{X \subset \Theta \text{ or } Y \subset \Theta, X \cap Y \subseteq A} m_1(X)m_2(Y)}{\sum_{X \cap Y \neq \emptyset} m_1(X)m_2(Y)}$$

for all $A \neq \emptyset, \Theta$.

More generally, let $m_1, m_2, ..., m_n; n > 1$ be mass functions on $2^\Theta$ and let $pls_1, pls_2, ..., pls_n$ be their plausibility functions, respectively. Let $0 < \alpha < 1$. Then we have the following formula for $(pls_1 \oplus pls_2 \oplus ... \oplus pls_n)^\alpha$:

$$(pls_1 \oplus pls_2 \oplus ... \oplus pls_n)^\alpha(\emptyset) = 0,$$
$$(pls_1 \oplus pls_2 \oplus ... \oplus pls_n)^\alpha(\Theta) = 1,$$
$$(pls_1 \oplus pls_2 \oplus ... \oplus pls_n)^\alpha(A) = 1 - (1-\alpha) \times$$

$$\times (\sum_{X_1 \subset \Theta \text{ or } ... \text{ or } X_n \subset \Theta, X_1 \cap X_2 \cap ... \cap X_n \subseteq A} m_1(X_1)m_2(X_2)...m_n(X_n))$$

$$\div (\sum_{X_1 \cap X_2 \cap ... \cap X_n \neq \emptyset} m_1(X_1)m_2(X_2)...m_n(X_n))$$

for all $A \neq \emptyset, \Theta$.

**THEOREM 8.** (*Combining discounted plausibility functions.*) Let $m_1, m_2$ be mass functions on $2^\Theta$ and let $pls_1, pls_2$ be their plausibility functions, respectively. Let $0 < \alpha < 1$. Then we have the following formula for $pls_1^\alpha \oplus pls_2^\alpha$: $(pls_1^\alpha \oplus pls_2^\alpha)(\emptyset) = 0, (pls_1^\alpha \oplus pls_2^\alpha)(\Theta) = 1$, and

$$(pls_1^\alpha \oplus pls_2^\alpha)(A) = 1 - ((1-\alpha)^2 (\sum_{X \cap Y \subseteq A} m_1(X)m_2(Y))$$

$$+ \alpha(1-\alpha)(\sum_{Z \subseteq A}(m_1(Z) + m_2(Z))))$$

$$\div ((1-\alpha)^2 \sum_{X \cap Y \neq \emptyset} m_1(X)m_2(Y) + 2\alpha - \alpha^2)$$

for all $A \neq \emptyset, \Theta$.

### 3.5 THE FORMULAE FOR $(com_1 \oplus com_2)^\alpha$ AND $com_1^\alpha \oplus com_2^\alpha$

Next, commonality functions.

**THEOREM 9.** (*Discounting an orthogonal sum of commonality functions.*) Let $m_1, m_2$ be mass functions on $2^\Theta$ and let $com_1, com_2$ be their commonality functions, respectively. Let $0 < \alpha < 1$. Then we have the following formula for $(com_1 \oplus com_2)^\alpha$: $(com_1 \oplus com_2)^\alpha(\emptyset) = 1$,

$$(com_1 \oplus com_2)^\alpha(\Theta)$$
$$= (1-\alpha)\frac{m_1(\Theta)m_2(\Theta)}{\sum_{X \cap Y \neq \emptyset} m_1(X)m_2(Y)} + \alpha,$$

and

$$(com_1 \oplus com_2)^\alpha(A)$$
$$= (1-\alpha)\frac{\sum_{\Theta \supseteq X \cap Y \supseteq A} m_1(X)m_2(Y)}{\sum_{X \cap Y \neq \emptyset} m_1(X)m_2(Y)} + \alpha$$

for all $\emptyset \subset A \subset \Theta$.

More generally, let $m_1, m_2, ..., m_n; n > 1$ be mass functions on $2^\Theta$ and let $com_1, com_2, ..., com_n$ be their commonality functions, respectively. Let $0 < \alpha < 1$. Then we have the following formula for $(com_1 \oplus com_2 \oplus ... \oplus com_n)^\alpha$:

$$(com_1 \oplus com_2 \oplus ... \oplus com_n)^\alpha(\emptyset) = 1,$$

$$(com_1 \oplus com_2 \oplus ... \oplus com_n)^\alpha(\Theta) = (1-\alpha) \times$$

$$\times \frac{m_1(\Theta)m_2(\Theta)...m_n(\Theta)}{\sum_{X_1 \cap X_2 \cap ... \cap X_n \neq \emptyset} m_1(X_1)m_2(X_2)...m_n(X_n)} + \alpha,$$

and $(com_1 \oplus com_2 \oplus ... \oplus com_n)^\alpha(A) = (1-\alpha) \times$

$$\times \frac{\sum_{\Theta \supseteq X_1 \cap X_2 \cap ... \cap X_n \supseteq A} m_1(X_1)m_2(X_2)...m_n(X_n)}{\sum_{X_1 \cap X_2 \cap ... \cap X_n \neq \emptyset} m_1(X_1)m_2(X_2)...m_n(X_n)} + \alpha$$

for all $\emptyset \subset A \subset \Theta$.

**THEOREM 10.** (*Combining discounted commonality functions.*) Let $m_1, m_2$ be mass functions on $2^\Theta$ and let $com_1, com_2$ be their commonality functions, respectively. Let $0 < \alpha < 1$. Then we have the following formula for $com_1^\alpha \oplus com_2^\alpha$: $(com_1^\alpha \oplus com_2^\alpha)(\emptyset) = 1$, $(com_1^\alpha \oplus com_2^\alpha)(\Theta)$

$$= ((1-\alpha)^2 m_1(\Theta)m_2(\Theta)$$
$$+ \alpha(1-\alpha)(m_1(\Theta) + m_2(\Theta)) + \alpha^2)$$
$$\div ((1-\alpha)^2 \sum_{X \cap Y \neq \emptyset} m_1(X)m_2(Y) + 2\alpha - \alpha^2),$$

and $(com_1^\alpha \oplus com_2^\alpha)(A)$

$$= \frac{(1-\alpha)^2 (\sum_{\Theta \supseteq X \cap Y \supseteq A} m_1(X)m_2(Y)) + \alpha^2}{(1-\alpha)^2 \sum_{X \cap Y \neq \emptyset} m_1(X)m_2(Y) + 2\alpha - \alpha^2}$$
$$+ \frac{\alpha(1-\alpha)(\sum_{\Theta \supseteq Z \supseteq A}(m_1(Z) + m_2(Z)))}{(1-\alpha)^2 \sum_{X \cap Y \neq \emptyset} m_1(X)m_2(Y) + 2\alpha - \alpha^2}$$

for all $\emptyset \subset A \subset \Theta$.

### 3.6 THE FORMULAE FOR $(dou_1 \oplus dou_2)^\alpha$ AND $dou_1^\alpha \oplus dou_2^\alpha$

Finally, doubt functions.

**THEOREM 11.** (*Discounting an orthogonal sum of doubt functions.*) Let $m_1, m_2$ be mass functions on $2^\Theta$ and let $dou_1, dou_2$ be their doubt functions, respectively. Let $0 < \alpha < 1$. Then we have the following formula for $(dou_1 \oplus dou_2)^\alpha$: $(dou_1 \oplus dou_2)^\alpha(\emptyset) = 1$, $(dou_1 \oplus dou_2)^\alpha(\Theta) = 0$,

$$(dou_1 \oplus dou_2)^\alpha(A)$$
$$= (1-\alpha)\frac{\sum_{X \subset \Theta \text{ or } Y \subset \Theta, X \cap Y \subseteq \bar{A}} m_1(X)m_2(Y)}{\sum_{X \cap Y \neq \emptyset} m_1(X)m_2(Y)}$$

for all $A \neq \emptyset, \Theta$.

More generally, let $m_1, m_2, ..., m_n; n > 1$ be mass functions on $2^\Theta$ and let $dou_1, dou_2, ..., dou_n$ be their doubt functions, respectively. Let $0 < \alpha < 1$. Then we have the following formula for $(dou_1 \oplus dou_2 \oplus ... \oplus dou_n)^\alpha$:

$$(dou_1 \oplus dou_2 \oplus ... \oplus dou_n)^\alpha(\emptyset) = 0, (dou_1 \oplus dou_2 \oplus ... \oplus dou_n)^\alpha(\Theta) = 1, (dou_1 \oplus dou_2 \oplus ... \oplus dou_n)^\alpha(A)$$

$$= (1-\alpha) \times (\sum_{X_1 \subset \Theta \text{ or } ... \text{ or } X_n \subset \Theta, X_1 \cap X_2 \cap ... \cap X_n \subseteq \bar{A}} m_1(X_1)m_2(X_2)...m_n(X_n))$$
$$\div (\sum_{X_1 \cap X_2 \cap ... \cap X_n \neq \emptyset} m_1(X_1)m_2(X_2)...m_n(X_n))$$

for all $A \neq \emptyset, \Theta$.

**THEOREM 12.** (*Combining discounted doubt functions.*)

Let $m_1, m_2$ be mass functions on $2^\Theta$ and let $dou_1, dou_2$ be their doubt functions, respectively. Let $0 < \alpha < 1$. Then we have the following formula for $dou_1^\alpha \oplus dou_2^\alpha$: $(dou_1^\alpha \oplus dou_2^\alpha)(\emptyset) = 1, (dou_1^\alpha \oplus dou_2^\alpha)(\Theta) = 0$, and $(dou_1^\alpha \oplus dou_2^\alpha)(A)$

$$= ((1-\alpha)^2 (\sum_{X \cap Y \subseteq \bar{A}} m_1(X)m_2(Y))$$
$$+ \alpha(1-\alpha)(\sum_{Z \subseteq \bar{A}}(m_1(Z) + m_2(Z))))$$
$$\div ((1-\alpha)^2 \sum_{X \cap Y \neq \emptyset} m_1(X)m_2(Y) + 2\alpha - \alpha^2)$$

for all $A \neq \emptyset, \Theta$.

## 4  SUMMARY

We have discussed how to discount a mass function and its belief function, plausibility function, commonality function, and doubt function to a support functions of the respective kinds.

We have also investigated the relation between the discount operation and the combination operation on evidential functions. In each case these two basic operations are non-commutative. We derived formulae for the combination of discounted evidential functions.